\documentclass[runningheads]{llncs}
\usepackage{wrapfig,lipsum,booktabs}
\usepackage{graphicx}
\usepackage{epsfig}
\usepackage{svg}
\usepackage{eucal}
\usepackage{booktabs}
\usepackage{todonotes}
\usepackage{verbatim}
\usepackage{amsmath}
\usepackage{amssymb}
\usepackage[normalem]{ulem}
\usepackage{colortbl}
\useunder{\uline}{\ul}{}

\usepackage{multirow}
\usepackage{todonotes}

\newcommand{\XX}{\mathbf{X}}
\newcommand{\UU}{\mathbf{U}}
\newcommand{\YY}{\mathbf{Y}}

\newcommand{\dd}{\mathbf{d}}
\newcommand{\hh}{\mathbf{h}}
\newcommand{\pp}{\mathbf{p}}
\newcommand{\qq}{\mathbf{q}}
\newcommand{\PP}{\mathbf{P}}

\newcommand{\real}{\mathbb{R}}

\begin{document}

% \title{Segmentation Adaptation without the source}
\title{On the pitfalls of entropy-based uncertainty for multi-class semi-supervised segmentation}
%for~Image~Segmentation}

%\title[Constrained Domain Adaptation for Segmentation]{Constrained Domain Adaptation for Segmentation}

% commenting out to put the anonymous section (avoiding desk-rejection from space cheats)
 %\author{******* **********, ***** ******, ******** *******, ********* ********}
 
  \author{Martin Van Waerebeke$^1$, Gregory Lodygensky$^2$, Jose Dolz$^3$}
 \institute{$^1$CentraleSupélec, Paris, France\footnote{Work done as part of a research internship at ETS Montreal and CHU Sainte-Justine.}\\$^2$CHU Sainte-Justine, Montreal, Canada\\
 $^3$ETS Montreal, Montreal, Canada}
% More complicate cases, e.g. with dual affiliations and joint authorship

%

\maketitle

\newcommand{\mathbbm}[1]{\text{\usefont{U}{bbm}{m}{n}#1}}

%\begin{document}

%\maketitle

\begin{abstract}

Semi-supervised learning has emerged as an appealing strategy to train deep models with limited supervision. Most prior literature under this learning paradigm resorts to dual-based architectures, typically composed of a teacher-student duple. To drive the learning of the student, many of these models leverage the aleatoric uncertainty derived from the entropy of the predictions. While this has shown to work well in a binary scenario, we demonstrate in this work that this strategy leads to suboptimal results in a multi-class context, a more realistic and challenging setting. We argue, indeed, that these approaches underperform due to the erroneous uncertainty approximations in the presence of inter-class overlap. Furthermore, we propose an alternative solution to compute the uncertainty in a multi-class setting, based on divergence distances and which account for inter-class overlap. We evaluate the proposed solution on a challenging multi-class segmentation dataset and in two well-known uncertainty-based segmentation methods. The reported results demonstrate that by simply replacing the mechanism used to compute the uncertainty, our proposed solution brings substantial improvement on tested setups.

\end{abstract}


\section{Introduction}

Semantic segmentation on medical images plays an essential role in a variety of clinical applications, including diagnosis, treatment planning or personalised medicine. Modern deep neural networks are driving progress in this task, demonstrating an astonishing performance when large labeled datasets are available. Nevertheless, obtaining such curated pixel-level masks is a highly time-consuming task, prone to inter-observer variability, which is further magnified when volumetric images are involved.

To alleviate the need for such large labeled datasets, several learning strategies have recently emerged as appealing alternatives to fully supervised training. For example, weak supervision, which can come in the form of image-tags \cite{kervadec2019constrained,patel2022weakly}, scribbles \cite{valvano2021learning} or bounding boxes \cite{KervadecMiccai}, is paving the way towards closing the gap between weakly supervised and fully supervised models. Even though this setting reduces the burden of the labeling process, achieved performances are typically far from their fully supervised counterparts. In contrast, semi-supervised learning (SSL), which learns from a limited amount of labeled data and a large set of unlabeled data, offers an interesting balance between high-quality segmentation and low labeled data regime.

Despite the different nature of the many existing approaches, the prevalent principle of these techniques is to augment a standard supervised loss --on a reduced number of labeled images--, with an unsupervised regularization term using unlabeled images. Adversarial learning based methods \cite{fang2020dmnet,zhang2017deep} typically leverage a discriminator network to encourage the segmentation model to provide similar segmentation outputs for both labeled and unlabeled images. Co-training \cite{blum1998combining} leverages the underlying idea that training examples can often be described by two complementary sets of features, often referred to as views, mining consensus information from multiple views \cite{zhou2019semi,xia20203d,peng2020deep,wang2021self}. More recently, contrastive-learning \cite{hadsell2006dimensionality} has also been adopted in the context of semi-supervised medical image segmentation \cite{chaitanya2020contrastive,peng2021self}. In particular, the model is pre-trained in an unsupervised manner with a contrastive loss, improving its performance for a subsequent downstream task, for which a few labeled samples are available.

Nevertheless, despite this variety of methods, consistency regularization is emerging as an appealing alternative to existing approaches. The common idea behind these methods is to have a dual-based architecture, similar to the teacher-student duple, and enforce consistency between the predictions of the same image under a given set of perturbations. Input modifications can include affine transformations \cite{li2020transformation}, intensity modifications \cite{xu2021shadow} or image permutations \cite{li2020self}, among others. In addition, \cite{luo2021semi} integrated a regression task to generate signed distance maps of the corresponding target object from the input image, which were employed to enforce the consistency with the pixel-level predicted masks. To account for noisy labels and errors produced in the pseudo-segmentation masks, %which might result in a significant degradation in performance, % used in the consistency terms, 
most recent papers resort to uncertainty-based approaches, which weight the reliability of each pixel in the objective function. By integrating this uncertainty-awareness scheme, the student model can gradually learn from reliable targets by exploiting the uncertainty information, ultimately resulting in better segmentation results. More concretely, the predictive entropy has been commonly preferred to approximate the uncertainty, as originally proposed in \cite{yu2019uncertainty}. In particular, authors resort to Monte Carlo dropout and random Gaussian noise to generate multiple segmentations from the same image, even though other strategies, such as temporal ensembling, can also be employed. This approach has been further improved by integrating additional components \cite{wang2020double,yang2020deep,wang2021tripled,zhang2021uncertainty,hu2022semi}. For example, \cite{wang2020double} explores a dual uncertainty method which, in addition to the predictive uncertainty, includes the uncertainty derived from the learned features. However, from all of the papers we have read leveraging predictive entropy as uncertainty, none has questioned its validity.

In this work, we challenge the status-quo of SSL methods resorting to entropy-based uncertainty and demonstrate that, under the multi-class scenario, the results derived by these approaches are suboptimal. The proposed work is based on the observation that the entropy-based metric, such as the ones employed to compute the uncertainty in most SSL segmentation approaches, fail to account for any overlap between the output distribution of several %competing
classes. To illustrate this point, we provide a counterexample in Figure \ref{fig:entropy-flaws}. Furthermore, we experimentally evaluate the use of entropy as a measure of uncertainty for semi-supervised segmentation methods and demonstrate that it leads to suboptimal results in multi-class settings. In particular, we resort to two well-known methods that fall in this category, i.e., \cite{yu2019uncertainty} and \cite{wang2020double}, which are evaluated on a public multi-class segmentation benchmark. We find that, surprisingly, \textit{in the presence of multiple classes they bring none or marginal performance} compared to the baseline not integrating uncertainty. In contrast, by approximating the uncertainty with properly suited divergences, the problematic inter-class overlap is taken into account, ultimately resulting in better segmentation performances. We stress that we claim no novelty regarding a new SSL framework. Instead, we highlight the flaws of the current status-quo of existing entropy-based methods in the multi-class scenario and propose alternative metrics to address them. %\textcolor{red}{Note that most literature in SSL, particularly using entropy-based uncertainty, is empirically validated on binary settings, which further supports our findings. }

\section{Methodology}

\textbf{Notation.} We denote the set of labeled training images as $\mathcal{D_L} = \{(\XX_n, \YY_n)\}_n$, where $\XX_i \in \real^{\Omega_i}$ represents the \textit{i}$^{th}$ image, $\Omega_i$ its spatial domain and $\YY_i \in \{ 0,1 \}^{\Omega_i \times C}$ its corresponding ground-truth segmentation mask. Furthermore, the set of unlabeled images is denoted as $\mathcal{D_U} = \{(\XX_m)\}_m$, which contains only images and where $m >> n$. The goal of semi-supervised semantic segmentation is to learn a segmentation network $f(\cdot)$, by leveraging both the labeled and unlabeled datasets. Note that in dual-based architectures, there exist two different networks: the teacher, parameterized by $\theta_T$ and the student, represented by $\theta_S$. In addition, for each image $\XX_i$, we denote $\PP_i \in [ 0,1 ]^{\Omega_i \times C}$ as the softmax probability output of the network, i.e., the matrix containing a simplex column vector $\pp_i^v = \left ( p_i^{v,1}, \dots, p_i^{v,C} \right )^{T} \in [0, 1]^C$ for each voxel $v \in \Omega_i$. Note that we omit the parameters of the network here to simplify notation. 

\subsection{Background}

In this section, we revisit the standard setting for SSL segmentation based on uncertainty, which aim at minimizing the following combined objective:

\begin{equation}
    \min_\theta \sum_{i=1}^N \mathcal{L}_s (f_{\theta_S}(\XX_i),\YY_i) + \lambda \sum_{i=1}^{N+M} \mathcal{L}_c (f_{\theta_T}(\XX_i; \xi^T), f_{\theta_S}(\XX_i; \xi^S))
    \label{eq:ssl}
\end{equation}

where $\mathcal{L}_s$ can be any segmentation loss on the labeled images, and $\mathcal{L}_c$ is a consistency loss to enforce the similarity between teacher and student predictions for the same input $\XX_i$ under different perturbations\footnote{Adding noise to input images and dropout are standard perturbations in SSL.}, $\xi^T$ and $\xi^S$. Furthermore, while the student parameters are updated via standard gradient descent, the teacher weights are updated as an exponential moving average (EMA) \cite{tarvainen2017mean} of the student parameters at different time steps. To account for unreliable and noisy predictions made by the teacher in low-data regimes, \cite{yu2019uncertainty} introduced an uncertainty-aware consistency loss, which has been employed by many more recent works. In particular, for a given image $\XX_i$, this loss can be defined as:

\begin{equation}
    \mathcal{L}_c (\PP_{\theta_T}, \PP_{\theta_S})=\frac{\sum_{v \in \Omega_i} \mathbb{I}(u_v < H) \| \PP_{\theta_T, v} - \PP_{\theta_S, v} \|^2 }{\sum_{v \in \Omega_i} \mathbb{I}(u_v < H)}
\end{equation}

where $\mathbb{I}$ is the indicator function, $H$ is a threshold to select the most certain voxels and $u_v$ is the estimated uncertainty at voxel $v$. 

\textbf{Uncertainty estimation.} Epistemic uncertainty is modeled using Monte Carlo DropOut (MCDO) \cite{gal2016dropout,kendall2017uncertainties}. More concretely, $T$ stochastic forward passes are typically performed on the teacher model, each with random dropout and Gaussian noise, which results in a set of softmax probability predictions per volume, $\{\PP^t\}_{t=1}^T$. In the current semi-supervised segmentation literature, the predictive entropy is preferred to approximate these uncertainty estimates. Thus, for each voxel $v$ in the \textit{i-th} volume, we can compute the average softmax prediction for class $c$ as $\mu_c=\frac{1}{T}\sum_t \pp^c_t$. Finally, the voxel-wise uncertainty is computed as $u^v=-\sum_c \mu_c \log \mu_c$.

\begin{figure}[h!]
    \centering
    \vspace{-4 mm}
    \includegraphics[width=1.0\textwidth]{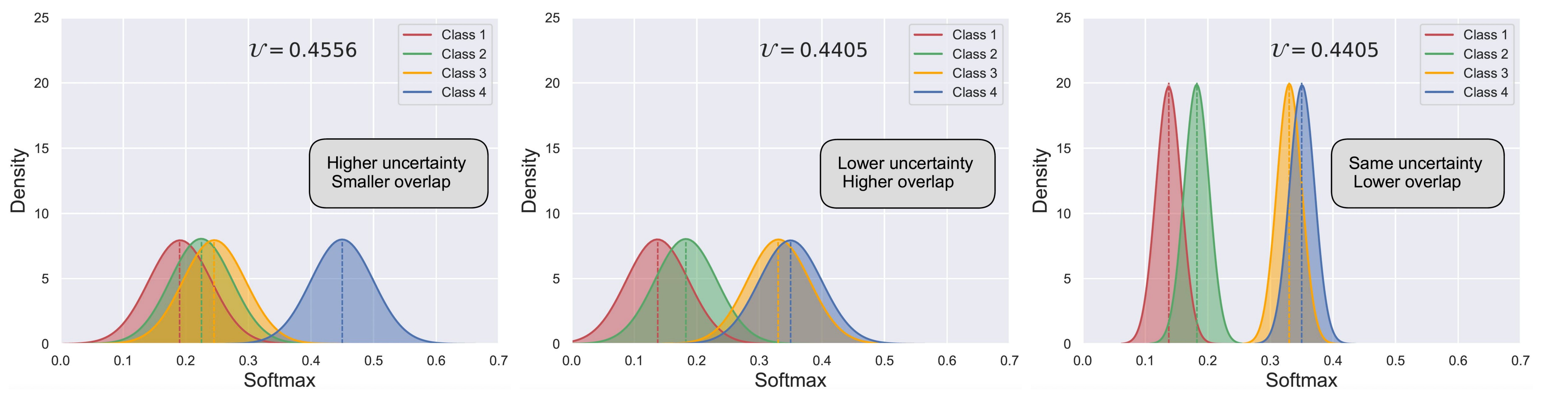}
    \caption{Counterexample that highlights the weaknesses of entropy to approximate the uncertainty. Class prediction distributions and their corresponding uncertainty value.}%, are reported.} %Barplots in the \textit{top} represent the per-class softmax predictions for the same sample under two different perturbations with \textit{high uncertainty} (first and second plots) and \textit{low uncertainty} (third and fourth), according to variance-based uncertainty. In the \textit{bottom}, we depict the per-class predictive distribution. }
    \vspace{-10 mm}
    \label{fig:entropy-flaws}
\end{figure}

\subsection{Limitations of entropy-based uncertainty}

Recent literature has been systematically integrating the predictive entropy derived from multiple predictions as an approximation of the uncertainty, without questioning its validity. We challenge these standard practices and argue that entropy-based metrics, such as the ones employed in the existing literature, yield suboptimal results in multi-class scenarios. Our hypothesis is based on the observation that entropy-based metrics ignore the inter-class separability, overlooking potentially high overlapping between the distributions of softmax outputs over multiple perturbations for competing classes. We demonstrate this negative effect with a counterexample in Figure \ref{fig:entropy-flaws}. These plots show the class probability distributions of a given sample under different perturbations\footnote{Note that in our setting, a sample represents a single voxel, which may result in different softmax predictions due to the Monte Carlo dropout step.}. In particular, we depict three different scenarios, where the class 4 is most likely to be predicted. For each setting, we show the predictive distributions across classes and their predicted entropy-based uncertainty. In the first case (\textit{left}), we can observe that the distribution of the predictions results in almost no overlap between the distribution of the winner class (i.e., class 4) and the rest of the classes. In this case, one would safely argue that this sample belongs indeed to the class 4. This contrasts with the estimated uncertainty, which flagged the prediction as more uncertain than its counterpart in the middle plot. On the other extreme, we can observe that predicted distributions for classes 3 and 4 highly overlap. Nevertheless, this strong overlap between the two classes is not taken into account to compute the uncertainty, whose obtained value is lower than in the left case. In other words, the entropy-based uncertainty has identified the middle example as having lower uncertainty, whereas one can clearly see that choosing between class 3 or 4 cannot be done confidently. Last, the third plot highlights another limitation of this type of uncertainty. Indeed, it fails to account for variance of the predictions, providing the same uncertainty value for the \textit{middle} and \textit{right} scenarios, whereas the standard deviation of the prediction distributions are wildly different. Even more, in the extreme case in which the variances were close to zero, but the means of the distributions remained unchanged, the predicted uncertainty values would stay the same.  This counterexample demonstrates that, in a multi-class setting, the use of entropy-based metrics to approximate the uncertainty produce suboptimal results, as the inter-class relations are not taken into account in the estimation of the uncertainty. Furthermore, as the vast majority of semi-supervised segmentation methods resorting to uncertainty have been evaluated only in the binary setting, these limitations have been largely overlooked.

\subsection{Alternatives to the Shannon entropy}
\label{ssec:our}

To overcome the limitations of approximating the uncertainty with predictive entropy in a multi-class scenario, we propose using divergence distances that account for inter-class overlaps. %\textcolor{red}{Thus, any solution to this issue must take into account the distribution of softmax probabilities, and not only their means}. 
To begin with, we draw inspiration from uncertainty based approaches and perform $T$ inferences per image, each of them with random dropout and Gaussian noise. Then, we compute the sample distributions for each class $C$ at every voxel $v$, resulting in the following distributions $\dd_1^v,...,\dd_C^v$. The distributions are sorted in increasing order based on their means. To measure the overlapping between different distributions, we build a set of histograms, one per distribution, resulting in $\hh_1^v, ..., \hh_C^v$.    
We must now evaluate how much each histogram stand out from the others. The most intuitive response is then to have a look at existing statistical distances.

\paragraph{\textbf{Bhattacharyya.}} We first resort to the Battacharyya divergence \cite{bhattacharyya1946some}, which has recently been proposed for class separability. More concretely, \cite{pandy2021transferability} proposed to quantify the transferability of deep models across domains by leveraging the class separability in the feature space via a Gaussian Bhattacharyya coefficient. In addition, \cite{van2021leveraging} explored a Bhattacharyya coefficient to better approximate the inter-class confusion in uncertainty estimation compared to variance-based metrics. For two discrete distributions with $K$ samples, $\pp=(p_k)^K_{k=1}$ and $\qq=(q_k)^K_{k=1}$, the Bhattacharyya divergence can be formally defined as:
\begin{equation}
\mathcal{D}_{BC}(\pp,\qq)= - \log \sum_{k=1}^K \left ( p_k q_k \right )^{\frac{1}{2}} 
\label{eq:bc}
\end{equation}

\paragraph{\textbf{Alpha-divergences.}}We also investigate the Tsallis’s formulation of $\alpha$-divergence \cite{cichocki2010families,tsallis1988possible,amari2009alpha,havrda1967quantification} which, by using the generalized logarithm \cite{cichocki2010families}: $ \log_\alpha(x)=\frac{1}{1-\alpha}(x^{1-\alpha}-1)$ extends the Kullback-Leibler (KL) divergence. Furthermore, with $\alpha=0.5$ and $\alpha=2.0$, it is equivalent to the Hellinger and Pearson Chi-square distances, respectively. The $\alpha$-divergence is defined as:

\begin{align}
\mathcal{D}_{\alpha}(\pp\|\qq) &= -\sum_{k=1}^K p_k \log_{\alpha} \left ( \frac{q_k}{p_k} \right) = \frac{1}{1-\alpha}  \left ( 1 - \sum_{k=1}^K p_k^{\alpha} q_k ^{1-\alpha}\right ) 
\label{eq:alphadiv}
\end{align}

Then, the voxel-wise uncertainty map $\UU$ can be derived from the pair-wise divergence distances as: $u_v= \max_{C>1} \mathcal{D}(\hh_1^v,\hh_C^v)$, where $\mathcal{D}$ denotes the divergence employed. We would like to highlight that \cite{pandy2021transferability} computed all the pair-wise divergences across all classes, whereas \cite{van2021leveraging} only resorted to the two classes with the highest means on their distributions. In our particular case, we empirically observed that exploiting the information from all the classes did not bring any significant improvement, while it considerably increased memory consumption. Thus, the final uncertainty map can be approximated as: $u_v= \mathcal{D}(\hh_{C-1}^v,\hh_C^v)$, where $\hh_{C-1}^v$ and $\hh_{C}^v$ are the histograms of the 2-top classes at voxel $v$. %In these metrics, high values represent high uncertainties, whereas low values indicate low uncertainties.

\section{Experiments}

%\subsection{Experimental Setting}

\paragraph{\textbf{Dataset.}} We perform our empirical validation on the publicly available developing Human Connectome Project (dHCP) dataset \cite{hughes2017developing}, which contains multi-modal MR images (MRI) from infants born at term. In particular, we employ 440 MRI-T2 images, and their %scanners acquired shortly after birth. %(37–44 weeks of gestational age). 
corresponding ground truth, which was generated using DrawEM and complemented further via manual correction. As per defined in the dataset, labels contain 9 classes: Cerebrospinal fluid (CSF), Cortical Gray Matter (CGM), Deep Gray Matter (DGM), White Matter (WM), Ventricles (V), Cerebellum (Cr), Hipoccampus and Amygdala (H-A), Brainsteam (Br) and Others (O). The dataset was divided into 3 labeled and 331 unlabeled images for training, 1 for validation and 105 for testing. Last, we crop the volumes into 64 $\times$ 64 $\times$ 64 patches and final segmentations were obtained using a sliding window strategy.

\paragraph{\textbf{Implementation details.}}We employ a 3D UNet as backbone architecture for all the models, with cross-entropy and Dice loss as objective function. The networks parameters were optimized with SGD (momentum=0.9) for a maximum of 200 epochs. The initial learning rate was set to $10^{-3}$ and divided by 2 every 15 epochs, with a batch size equal to 6. The EMA weighting factor was fixed at 0.99, %, following the literature. 
and $\lambda$ in equation (\ref{eq:ssl}) is set to 15 for all methods. Furthermore, we do not employ any data-augmentation to isolate the impact of both entropy-based and proposed uncertainty estimations. To generate multiple inferences per image, we follow standard literature (MCDO and Gaussian noise) with the same number of inferences $T$ across models, and use $10$ bins to compute the histograms. %Regarding the $\alpha$-divergence families, we evaluate their performance for $\alpha=0.5$ and $\alpha=2.0$, which are equivalent to the Hellinger and Pearson Chi-square distances, respectively.  

\paragraph{\textbf{Methods.}}To empirically validate our hypothesis, we conduct experiments on two well-known semi-supervised segmentation approaches: UA-MT \cite{yu2019uncertainty} and DU-MT \cite{wang2020double}. In addition to their popularity, another important factor is their simplicity, which facilitates the isolation of the entropy effect on the final performance. Furthermore, to have a better overview of the relative performance of these approaches, we include a \textit{lower baseline}, which is trained without additional unlabeled data, and an \textit{upper baseline}, trained on the whole labeled training set.

\paragraph{\textbf{Evaluation metrics.}}For evaluation purposes, we resort to common segmentation metrics in the medical imaging: Dice and Jaccard coefficients, the 95\% Hausdorff Distance (HD) and the average surface distance (ASD).

\subsection{Quantitative Results}

We first evaluate the effect of the entropy-based uncertainty on the segmentation performance of two well-known semi-supervised approaches (UA-MT and DU-MT). To this end, we report the per-class Dice scores obtained by the different baselines (Table \ref{table:main}). Furthermore, we replace the entropy as an approximator of the uncertainty by the Bhattacharyya divergence, denoted as \textit{ours}. From these results, we can observe that by resorting to a divergence distance, e.g., Bhattacharyya, the performance is considerably improved %($\sim2-2.5\%$ in terms of average dice) 
for both analyzed approaches. Furthermore, the improvement is consistent across all the classes, suggesting that the inter-class relationship is indeed better modeled.

\vspace{-2 mm}

\begin{table}[]
\scriptsize
\centering
\caption{Per-class dice scores obtained across different methods. Our approaches are shadowed, and $\nabla$ indicates the difference wrt the original (entropy-based) approach.}
\begin{tabular}{l|c|ccccccccc|c|c}
\toprule
& $\#N$ & CSF & CGM & WM & O & V & Cr & DGM & BS & H-A & Mean & $\nabla$ \\
\midrule
Lower-bound & 3 & 91.31 & 92.27 & 93.99 & 86.94 & 88.01  & 92.18 & 91.93 & 90.67 & 76.93 & 89.35 & --\\
\midrule
UA-MT \cite{yu2019uncertainty} & 3 & 90.84 & 91.91 & 93.66 & 85.93 & 88.60 & 93.57 & 92.43 & 92.23 & 78.35 &  89.72 & -- \\
\rowcolor{gray!20} UA-MT (ours) & 3 & 93.25 & 94.17 & 95.93 & 88.61 & 91.44 & 96.09 & 94.77 & 94.62 & 80.18 & 92.12 & +2.40
 \\
DU-MT \cite{wang2020double} & 3 & 90.74 & 91.74 & 93.46 & 85.89 & 88.16 & 91.37 & 90.88 & 90.23 & 73.73 & 88.47 & -- \\ 
\rowcolor{gray!20} DU-MT (ours) & 3 & 92.31 & 93.34  & 95.71 &  88.51 & 87.43 & 94.25 & 93.82 & 93.03 & 75.44 & 90.44 & +1.97\\
\midrule
Upper-bound & 334  & 96.42 & 97.51 & 98.37 & 92.85 & 95.92 & 98.09 & 97.91 & 97.98 & 94.20 & 96.58 \\
\bottomrule
\end{tabular}

\label{table:main}
\end{table}

In addition, Table \ref{table:all} reports the whole quantitative results across the different methods, which gives a better overview of their performance. We can see that the same trend is observed across the different metrics, with the proposed solution to approximate the uncertainty outperforming the original entropy-based methods. A surprising finding is that, compared to the lower-bound method, both original UA-MT and DU-MT \textit{bring none or marginal performance}. We argue that this is due to the highlighted limitation of the entropy as an estimation of uncertainty in the multi-class setting. Indeed, at the beginning of the training, the softmax probabilities for the different classes are closer, resulting in higher overlapped regions which might lead to unreliable uncertainty values.

\vspace{-3 mm}

\begin{table}[h!]
\scriptsize
\centering
\caption{Quantitative results for different segmentation metrics, where the mean over the 9 classes is reported.}
\begin{tabular}{l|c|cccc}
\toprule
& $\#N$ & Dice & Jaccard & HD95 & ASD \\
\midrule
Lower-bound & 3 & 89.35 & 82.23 & 3.39 & 1.65\\
\midrule
UA-MT \cite{yu2019uncertainty} & 3 &  89.72 & 81.77 & 3.20 & 1.14  \\
\rowcolor{gray!20} UA-MT (ours) & 3 &  92.12 & 85.89 & 2.32 & 0.67 \\
DU-MT \cite{wang2020double} & 3 &  88.47 & 79.96 & 4.10 & 1.33 \\ 
\rowcolor{gray!20} DU-MT (ours) & 3 &  90.44 & 83.16 & 2.87 & 1.04 \\
\midrule
Upper-bound & 334 & 96.58 & 93.48 & 1.19 & 0.28 \\
\bottomrule
\end{tabular}
\label{table:all}
\end{table}

\vspace{-5 mm}

\paragraph{\textbf{On the divergence distances.}} In this section we explore several divergence distances as alternative to the Bhattacharyya divergence. Results from this study, which are reported in Table \ref{table:divergences} demonstrate that, regardless of the divergence employed, the entropy-based uncertainty leads to sub-optimal results in both UA-MT and DU-MT approaches. Note that the balancing term in equation \ref{eq:ssl} remained fixed across all the models. However, we believe that further exploration of gradient magnitudes during training should be investigated to obtain optimal weighting parameters, particularly for Bhattacharyya and $\alpha-$divergences.

\vspace{-5 mm}

\begin{table}[h!]
\centering
\scriptsize
\caption{Ablation study on divergence distances, where the mean over the 9 classes is reported. Best results in bold, whereas second best results are underlined.}
\begin{tabular}{l|c|cccc}
\toprule
                        & Uncertainty & Dice & Jaccard & HD95 & ASD          \\ 
                        \midrule
\multirow{4}{*}{UA-MT \cite{yu2019uncertainty}}  &    Entropy& 89.72   &   81.77  & 3.20  &  1.14 \\
 & Bhattacharyya & \bf 92.12  &   \bf 85.89 &  \underline{2.32}     &   \bf 0.67               \\
 &  $\alpha$-div ($\alpha=0.5$) &   91.69 & \underline{85.15} & 2.37 & \underline{0.757} \\
  & $\alpha$-div ($\alpha=2.0$) & \underline{91.79} &   \underline{85.15}    & \bf 2.09 & 0.99\\
\midrule
\multirow{4}{*}{DU-MT \cite{wang2020double}}  &     Entropy& 88.47    &   79.96   & 4.1 & 1.32 \\
 & Bhattacharyya & 90.44  &     83.16   & 2.87 & 1.04         \\
 &  $\alpha$-div ($\alpha=0.5$) & \bf 90.99    &  \bf  84.01    & \bf 2.72 & \bf 0.92 \\
  &  $\alpha$-div ($\alpha=2.0$) & \underline{90.68}  &     \underline{83.64} & \underline{2.82}    & \bf 0.92 \\
\bottomrule
\end{tabular}
\label{table:divergences}
\end{table}

\vspace{-10 mm}

\subsection{Qualitative results}

\begin{figure}[h!]
    \centering
    \vspace{4 mm}
    \caption{Visual results of several scans in the axial view. In particular, ground truth, UA-MT with entropy and Bhattacharyya divergence, and DU-MT with entropy and Bhattacharyya divergence are shown. Green arrows highlight several differences between the entropy-based uncertainty method and its Bhattacharyya based counterpart. }
    \includegraphics[width=1.0\textwidth]{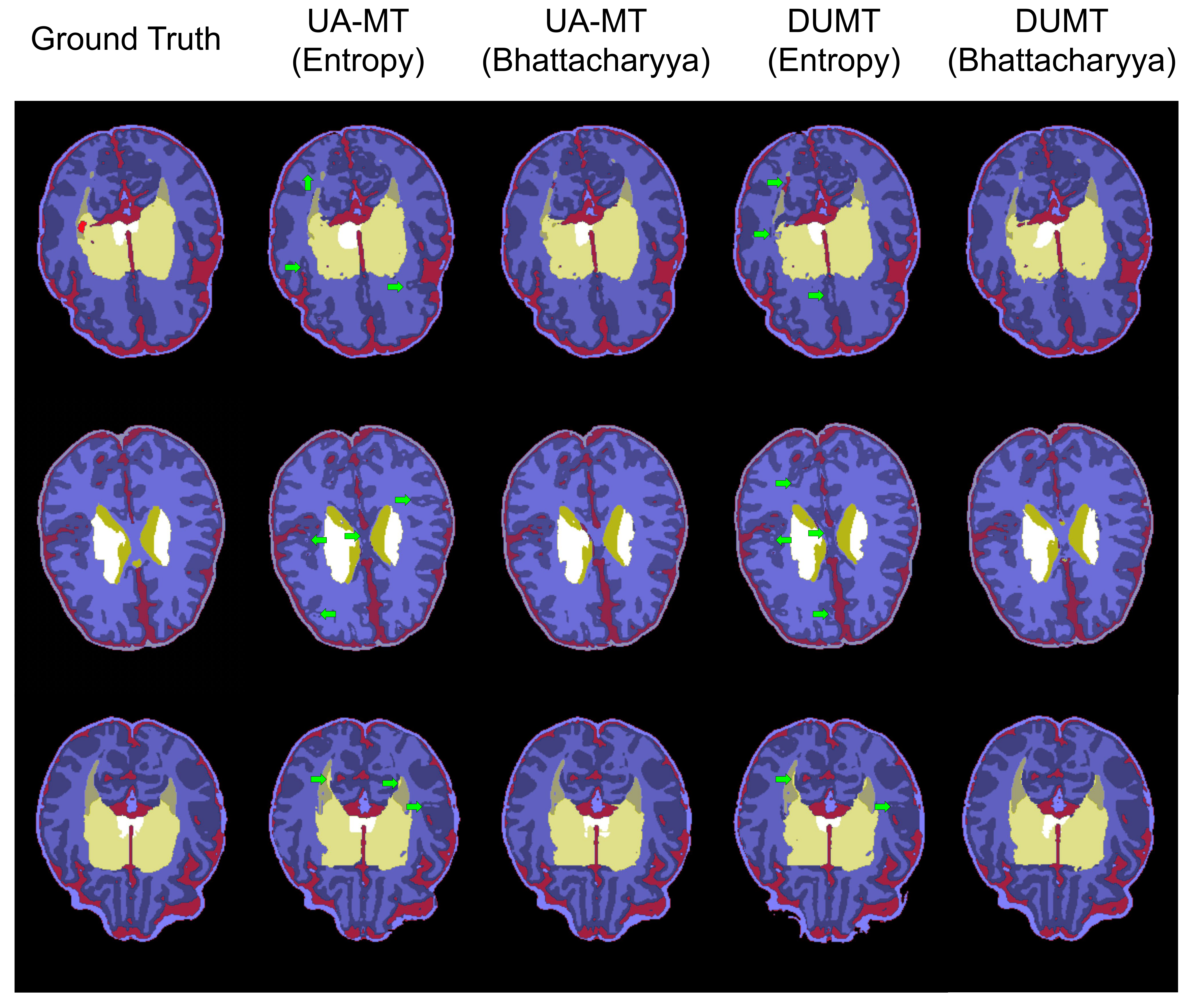}
    
    \label{fig:visual}
\end{figure}

We now present some qualitative results, which support the quantitative evaluation performed in previous sections. More concretely, in Figure \ref{fig:visual} we depict the effect of replacing the entropy-based uncertainty by the Bhattacharyya divergence. We can easily observe that the entropy-based uncertainty models systematically perform the same mistakes. For example, segmentations obtained by both UA-MT and DUMT approaches contain many holes in the gray matter. This confusion between gray and white matter is further magnified in the second row, where both methods identify all the regions at the middle of the hemispheres as gray matter, whereas they represent the white matter. In contrast, if we use a divergence distance instead, these mistakes are significantly corrected. We have employed green arrows to stress the location of several of these errors. 

\vspace{-10 mm}

\begin{figure}[h!]
    \centering
    \caption{Visual results of several scans in the sagittal view. In particular, ground truth, UA-MT with entropy and Bhattacharyya divergence, and DU-MT with entropy and Bhattacharyya divergence are shown. Green arrows highlight several differences between the entropy-based uncertainty method and its Bhattacharyya based counterpart. Please note the large differences particularly in the region of the corpus callosum.}
    \includegraphics[width=1.0\textwidth]{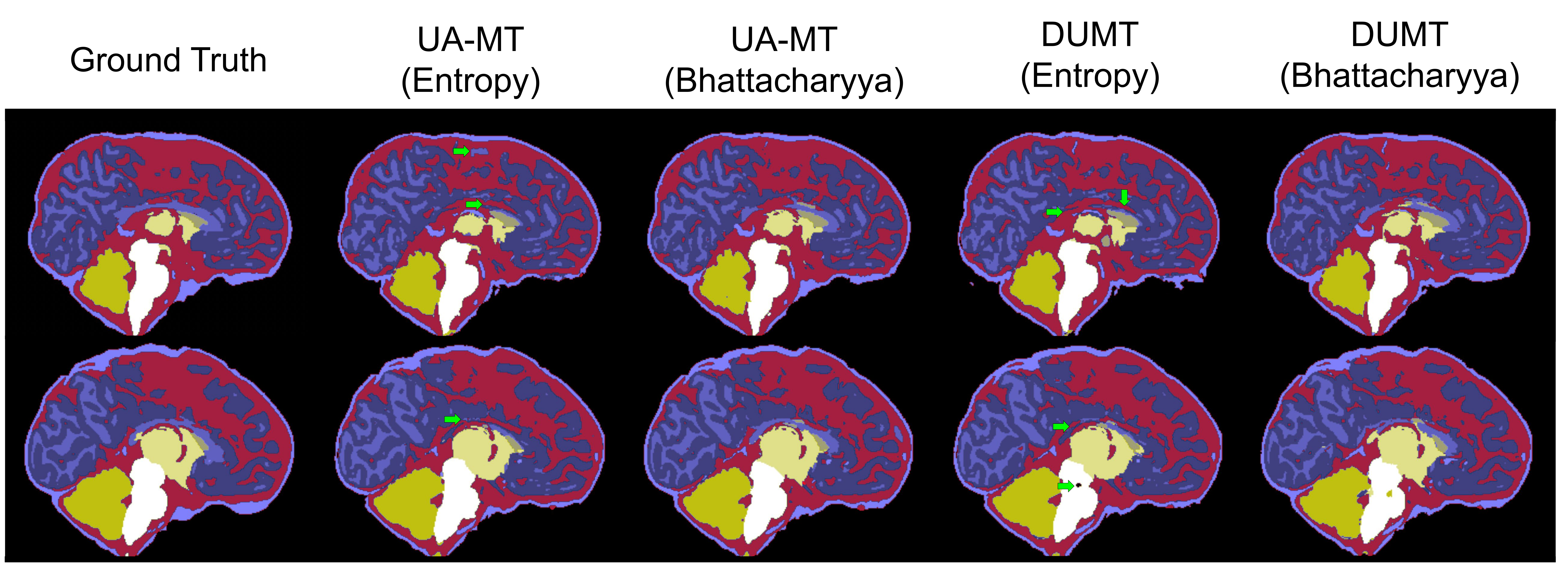}
    
    \label{fig:visual2}
\end{figure}

Furthermore, we also depict in Figure \ref{fig:visual2} two qualitative results on the sagittal view, where the lower performance of entropy-based uncertainty methods is better evidenced. In addition to the holes present in the gray matter, these approaches totally fail to correctly identify the region of the corpus callosum. Note, however, that despite there also exist differences in the segmentation results between our approaches and the ground truth, these are smaller compared to the baselines UA-MT and DUMT models.

\section{Discussion} 

In this work we have demonstrated an important limitation of entropy-based uncertainty for semi-supervised segmentation in the multi-class scenario. We have done this through an intuitive counterexample, which stressed the unreliable uncertainty estimates due to the erroneous approximations in the presence of several competing classes. We have further leveraged this finding to propose the use of several divergence distances as efficient alternatives for this task. The empirical validation on a challenging multi-class segmentation problem has supported our arguments related to the weaknesses of entropy-based methods, while confirming the superiority of divergence distances.

\bibliographystyle{splncs04}
\bibliography{biblio}

\end{document}